\documentclass[letterpaper]{article} 
\def\year{2020}\relax
\usepackage{aaai20}  
\usepackage{times}  
\usepackage{helvet} 
\usepackage{courier}  
\usepackage[hyphens]{url}  
\usepackage{graphicx} 
\usepackage{graphicx}  
\usepackage{amsmath}
\usepackage{amssymb}
\usepackage{enumitem}

\title{A Knowledge Driven Approach to Adaptive Assistance Using Preference Reasoning and Explanation}
\author{Jason R. Wilson\\Franklin \& Marshall College\\Lancaster, Pennsylvania\\jrw@fandm.edu \And Leilani Gilpin\\Massachusetts Institute of Technology\\Cambridge, Massachusetts\\lgilpin@mit.edu \And Irina Rabkina\\Occidental College\\Los Angeles, California\\irabkina@oxy.edu}

\begin{document}

\maketitle

\begin{abstract}
There is a need for socially assistive robots (SARs) to provide
transparency in their behavior by explaining their reasoning. Additionally,
the reasoning and explanation should represent the user’s preferences and
goals.  
To work towards satisfying this need for interpretable
reasoning and representations, we propose the robot uses Analogical
Theory of Mind to infer what the user is trying to do and uses the
Hint Engine to find an appropriate assistance based on what the user
is trying to do.  If the user is unsure or confused, the robot
provides the user with an explanation, generated by the Explanation
Synthesizer.  The explanation helps the user understand what the robot
inferred about the user’s preferences and why the robot decided to
provide the assistance it gave.  A knowledge-driven approach provides
transparency to reasoning about preferences, assistance, and 
explanations, thereby facilitating the incorporation
of user feedback and allowing the robot to learn and adapt to the user. 
\end{abstract}

\section{Introduction}
Socially assistive robots (SARs) can aid humans in a variety of tasks.
One of the most compelling assistive tasks is in medication
management, where a SAR can instruct, record, and oversee a patient's
medication usage.  However, since this is a medical application, it is
important that a robot is robust, transparent, and open to user
feedback; especially for corrections.

However, SARs, like other social robots, are complex to understand.  Robots are built of many parts, with underlying language tools (e.g., for NLP, NLU, or NLG) that are  not inherently interpretable.  Therefore, SARs cannot
effectively communicate and collaborate with humans on tasks without \emph{explainability}.  This is
troublesome when the robot fails, or when the assistive application is
critical, like healthcare or medical applications.  In this paper, we
make two distinct contributions towards explainable SARs: (1) we
contribute a complex cognitive model for incorporating user feedback,
and (2) we show a proof-of-concept on a real-life medication case
study.  Our approach combines three components:
\begin{enumerate}
\item Preference Reasoning: The robot considers what it knows or can infer about the user's preferences, and how these affect possible actions.
\item Assistance: The robot interacts with the user and aids
them if deemed necessary.
\item Explanation: The robot explains its reasoning and validates its
recommendation and conclusion with a user.
\end{enumerate}
In this paper, we show the capabilities of a SAR with knowledge-driven
adaptive assistance.  We start with a detailed overview of the
medication sorting task.  We present the approach and initial results,
and conclude with future work, discussion, and a reiteration of the
contributions.

\section{Task Description}
\label{sec:task-description}
We consider a SAR that provides social assistance in a medication sorting task.  In this task, a person organizes a set of medications, vitamins, and other supplements.  Each has constraints, provided in the form of a prescription, doctor recommendation, or personal preference.  In the example we use in this paper, there is a vitamin to be taken each morning and a medication that needs to be taken prior to any physical activity. 

\begin{figure}[hb]
    \centering
    \includegraphics[width=0.8\columnwidth]{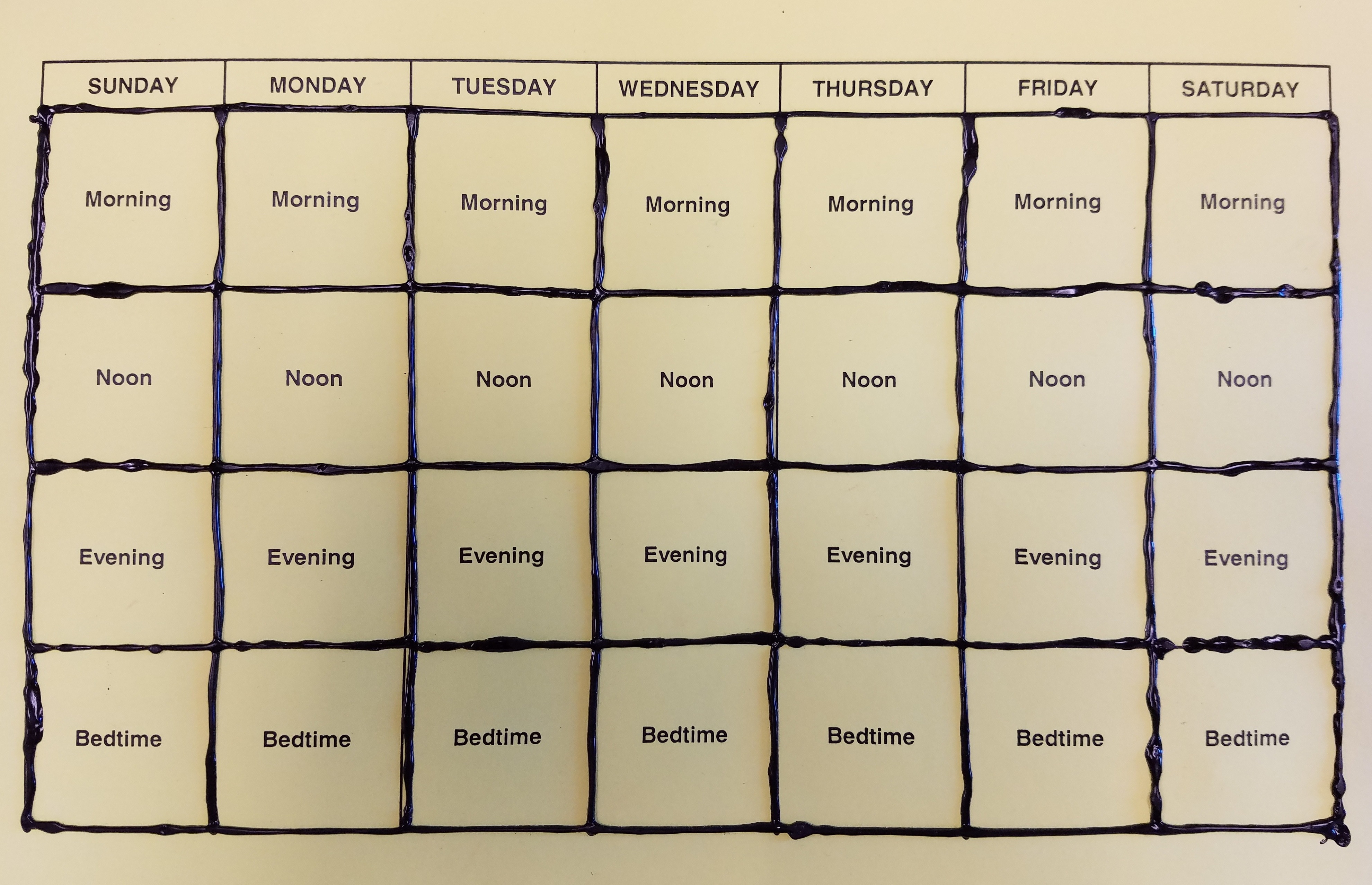}
    \caption{An example of a sorting grid.}
    \label{fig:grid}
\end{figure}

The role of the robot is to observe the person while they are organizing the pills into a sorting grid, a device that has compartments for each day of the week and multiple time of day (see Figure~\ref{fig:grid}).  Our example uses a sorting grid for four times in the day: morning, noon, evening, and bedtime.

Consider the scenario in which a person is to take a vitamin each morning and then Levodopa before any physical activity.  The robot knows that the person has a physical therapy appointment at 1pm on Wednesday and a dance class at 6pm on Friday.  The person has a preference that Levodopa is taken enough time before the activity for it to take effect. 

After the person has placed one vitamin in the morning compartment for each day, they begin to figure out where to place some Levodopa.  The person hesitates, and the robot interjects with a suggestion:

\begin{itemize}[left=30pt]
    \item[Robot:] Try placing a Levodopa pill in the morning on Wednesday. 
    \item[User:] Why?
    \item[Robot:] Levodopa needs to be taken before any physical activity, and you have a physical therapy appointment at 1pm on Wednesday.  Since you prefer to take it a few hours before activity, you should take it in the morning.
    \item[User:] Oh, right.  Thank you.
\end{itemize}

Alternatively, consider the case in which the person prefers to take Levodopa closer to when the activity is to occur.  In this case, the robot would explain:

\begin{itemize}[left=30pt]
    \item[Robot:] A Levodopa pill needs to be taken before any physical activity, and you have a physical therapy appointment at 1pm on Wednesday.  Since you prefer to take it immediately before activity, you should take it in the afternoon.
\end{itemize}

While previous work demonstrated the robot adapting its assistance based on how much assistance the person needed \cite{wilson2020challenges}, the current work looks at how the robot can reason about the person's preferences, thus adapting its assistance and explanation.

\section{Approach}

\begin{figure}[h]
\centering
\includegraphics[width=1.0\columnwidth]{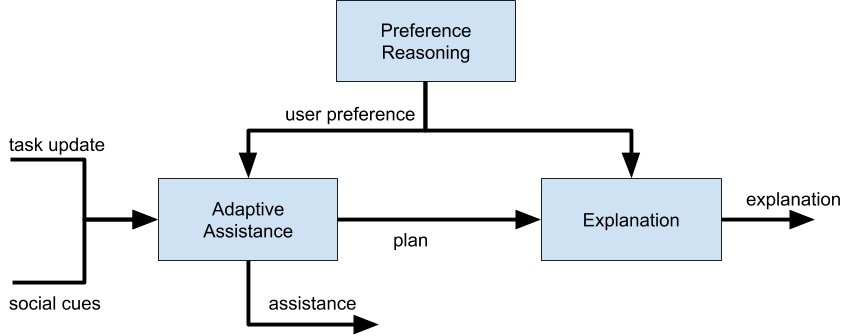} 
\caption{User preferences are sent to the Adaptive Assistance and Explanation components.  The Adaptive Assistance component uses the preference in generating a plan for completing the task, and the Explanation component uses the preference and plan to explain the robot's assistance.}
\label{fig:arch}
\end{figure}

For the social robot to provide assistance and explanations to the user, we propose an
architecture consisting of Preference Reasoning, Adaptive Assistance, and Explanation
components, as shown in Figure~\ref{fig:arch}.  The Adaptive Assistance component uses updates in the task and social cues conveyed by the user to generate a plan, which is used in determining what action the robot is to take to assist the user.  The plan would be sent to the Explanation component, along with the user preferences from the Preference Reasoning component, to generate an explanation for the robot's method of assistance. Each component is functional individually, but full integration is a work in progress

\subsection{Preference Reasoning}
In the context of pill sorting, user preferences can take two forms: preferences about how to sort specific pills (e.g., take Levodopa directly before an activity vs. several hours before) and preferences about the sorting task as a whole (e.g., sort all of one type of pill for the week vs. each day in order). These preferences take the form:
 \small{
 \begin{verbatim}
(prefers user 
    (medicationBeforeActivityBy 
        medtype
        distance))
 \end{verbatim}}
 \noindent 
 and
 
 \small{
 \begin{verbatim}(prefers user (sortOrder order))
  \end{verbatim}}
 \vspace{-4mm}
 \noindent 
 respectively. 
 
 Note that all preferences use the \texttt{prefers} predicate and take the user as the first argument. This representation allows us to generate preferences through both inference and user input. More importantly, it allows us to use the preferences for further reasoning, including in the Adaptive Assistance and Explanation components.
 
 In the present work, we assume that preferences are given by the user. This may be in the form of correcting the robot (e.g., "No, don't put that pill in the morning. I want to take it in the afternoon") or stated out right (e.g., "Let's start by sorting the green pills."). However, such preferences can also be inferred. We are working on integrating the Analogical Theory of Mind (AToM) ~\cite{rabkina2017towards} model into the architecture to do so.
 
 AToM is a computational cognitive model of the processes by which people learn to reason about others' preferences, goals, beliefs, desires, etc. (called theory of mind reasoning). It has successfully modeled children's learning in two developmental studies ~\cite{rabkina2017towards,rabkina2018bootstrapping} and has previously been used to recognize the goals and intentions of simulated agents in multiagent interactions ~\cite{rabkina2019analogical,rabkina2020acs}. 
 
 We plan to use AToM because its reasoning is human-like, and therefore easy for people to understand. Furthermore, AToM can learn from just a handful of examples and can incorporate user feedback to improve its reasoning on the fly.

\subsection{Adaptive Assistance}

The Adaptive Assistance component uses a Hint Engine to generate an appropriate assistance to help the user complete the task~\cite{wilson2018general}.  The Hint Engine integrates information from three models (need, assistance, and domain) to determine how and when the robot should assist \cite{wilson2019developing}.  The need model is used to infer how much assistance the person needs based on progress in the task and social cues (e.g., verbal requests, eye gaze patterns).  The assistance model represents the relations between the types of actions the user can take to complete the task, the types of actions the robot can take to assist, and the amount of assistance provided in any robot action.

The domain model represents the task that the user is performing and is used to infer a plan, the actions the user can take to complete the task.  When information from the need model indicates that the user has a sufficient level of need, the Hint Engine generates a plan for completing the task.  The first action in the plan indicates where the robot should focus its assistance.  Based on the action type and how much assistance the user needs, the Hint Engine uses the assistance model to determine the appropriate assistive action for the robot to perform.

\begin{figure}[h]
\centering
\includegraphics[width=0.99\columnwidth]{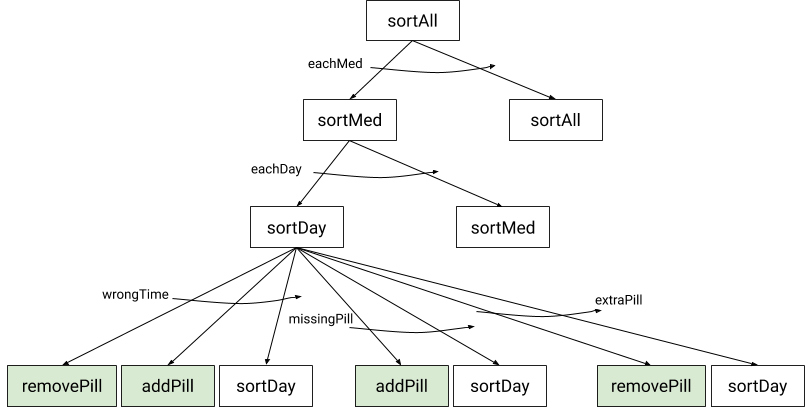} 
\caption{Hierarchical task network representing a medication sorting task. It assumes task is done by sorting all pills of one medication before doing the next one. White boxes are tasks, and shaded boxes are operators.  }
\label{fig:htn}
\end{figure}

To generate a plan, the Hint Engine uses its domain model, which is represented with a hierarchical task network (HTN) \cite{nau1999shop}.  The HTN used for medication sorting is shown in Figure~\ref{fig:htn}.  At the top level, the user is working toward sorting the pills for all of the medications.  To complete the task, the user needs to go through each medication, sorting the pills for each. For a given medication, a user may go through the days of the week, adding and removing pills as they go.  If no mistakes have been made, the user is just missing pills (leading to \texttt{addPill} actions).  If a pill has been placed at the wrong time of the day, then the action pair \texttt{removePill} and \texttt{addPill} are included in the plan to indicate the moving of a pill.

To determine whether there is a missing pill, an extra pill, or a pill placed at a wrong time, the planner 
checks the preconditions of the method, which includes the constraints defined for the given medication.  For example, each medication defines the maximum number to be taken in a day.  If the number of pills for that medication exceed the maximum, allowed, then the \texttt{extraPill} condition is satisfied.

We extend the precondition checks to also consider user preferences.  
For example, a vitamin could be taken at any time, but a user may prefer to take it in the morning.  Similarly, a medication like Levodopa might be taken before physical activity, and the user may have preferences regarding how long before the activity.  In this case, there is a constraint of the form \texttt{(beforeActivity pill row col activity)} that can be inferred with a rule like the following:

\small{
\begin{verbatim}
(beforeActivity pill row col activity) <-
    (activityAt activity rowX col)
    (isa pill med)
    (medicationBeforeActivityBy med 
        distance)
    (difference rowX row distance)
\end{verbatim}}

In this example, the HTN represents blocks of time as they relate to the sorting grid (i.e., morning, noon, etc.) and not specific times in the day.

\subsection{Explanation}
The explanations are generated from an existing
system \cite{gilpin2018monitoring,leilanithesis} that incorporates commonsense
knowledge, rules, and constraint checking towards an explanation of
intended behavior.  The Explanation Synthesizer proceeds in 3 steps:
\begin{enumerate}
\item Parsing and aggregation: The input query is parsed for key \emph{concepts}.  Those
concepts are used as search terms in the commonsense knowledge base.  A list of facts (symbolic triples) is returned.
\item Constraint Checking: Commonsense rules are triggered to generate
new facts and evidence.
\item Synthesizing: Once all the facts are aggregated, an explanation
synthesizer constructs the most plausible chain of reasoning
towards an explanation.
\end{enumerate}

For example, consider the query \texttt{(pill onDate Friday)} which justifies that the user can take the pill on Friday.  The query is parsed and
the key concepts are \texttt{pill} and \texttt{Friday}, which are search terms for
the commonsense knowledge base (KB).  The KB returns facts like
\texttt{(Friday IsA 'business day')}.  The relation \texttt{onDate} is
used as a constraint in the system.  

The constraints are a combination of commonsense rules and user
preferences.  These constraints are application dependent.   In the
medication sorting domain, they are rules related to the requirements
for each type of pill, as well as user preferences.  For example, the
user may prefer to take pills in the morning, or users may be
\emph{instructed} to ingest pills with meals or food.  The facts are
forward chained against these rules to generate new facts and
evidence.  

After this process, there may be more than one plausible explanation
supporting the query.  The explanation synthesizer starts from the
query and constructs a goal tree to satisfy the query
\cite{leilanithesis}.  For this paper, we only examine one
explanation.  Choosing the best explanation may be explored in future
work.

\section{Proof of Concept}


To demonstrate our components adapting to and explaining with user preferences, we consider the scenario in which the user has already correctly placed all of their Vitamin D and is now working to sort their Levodopa, which is to be taken before activity.  The user has two activities planned for the week, Wednesday at 1pm and Friday at 8pm.  The user just placed one Levodopa pill in the space for Monday midday.  

We assume that the user has a previously stated preference to take Levodopa during time slot prior to an activity.  When the user misplaces the Levodopa, the Adaptive Assistance component recognizes that the user needs assistance and generates a plan to 
move the Monday pill to an earlier time slot:

\small{
\begin{verbatim}
(planFor state8 
  ((preference beforeActivity 1)) 
  ((removePill Levodopa 3 1)
   (addPill Levodopa 3 0) 
   (addPill Levodopa 5 2)))
\end{verbatim}}

The first action, \texttt{(removePill Levodopa 3 1)}, is used along with an inference of a level of assistance to determine that the robot should provide direct assistance, which has the robot clearly stating what should be done next.  In this case, the pill needs to be removed and the robot would say ``Try removing a Levodopa from Wednesday''.

Additionally, the Adaptive Assistance component considers alternative plans, a counterfactual reasoning over different preferences.  The alternative plans do not affect how the robot assists but would be sent to the Explanation component.  An alternative plan for taking Levodopa at the same time as the activity is shown below. 
The plan indicates that the Monday pill is in the correct location and the only action remaining is to place the Friday pill:

\small{
\begin{verbatim}
(alternativePlanFor state8 
  ((preference beforeActivity 0)) 
  ((addPill Levodopa 5 3)))
\end{verbatim}}

Finally, the user can inquire about the robot's actions.  For example, the user can ask why the robot said to ``Try removing a Levodopa from Wednesday.''  This question is parsed into an intermediate representation: \texttt{(onDate Levodopa Wednesday)}, which is passed to the Explanation Synthesizer along with the associated \emph{preference}; the user prefers to take the medication before an activity.
The following is a trace of the reasoning of the Explanation Synthesizer

\small{
\begin{verbatim}
[(IsA Levodopa pill), 'Given']
[(AtLocation pill cabinet), 'ConceptNet']
...
[(IsA Wednesday weekday), 'ConceptNet']
[(IsA Wednesday day), 'ConceptNet']
...
[(prefers user (before pill activity)), 
    'Given preference']
[(IsA appt activity), 'Given knowledge']
[(atTime appt '1pm'), 'calendar']
[(onDay appt Wednesday), 'calendar']
[(atTime appt afternoon), 'Rule fired']
\end{verbatim}}

The justification is that \texttt{(onDay pill Wednesday) (beforeTime pill
afternoon)}.  And the final explanation reads in a series of symbolic
triples: \texttt{(prefers user (before pill activity)) (IsA user activity) (atTime appt '1pm') (onDay appt Wednesday) (IsA '1pm' afternoon)}.

\section{Future Work}
The work we describe here sets the foundation for a whole line of work in 
designing social robots to adapt to users, adhere to user preferences, and
provide explanations.  The most immediate next step is to build upon the
proof of concept we have demonstrated here by integrating the individual
parts and evaluating the system on more complex scenarios.

Once we have a fully integrated system, the critical next step is to 
incorporate feedback from the user.
One of the greatest advantages of taking a knowledge-driven approach is 
that the entire system is inspectable, which will facilitate integrating 
the user feedback.
The user may provide feedback after the robot provides an explanation to 
the user.  An explanation synthesizer extracts key terms from the
feedback, interprets the feedback, and determines which component(s)
need adjustment.  The explanation synthesizer also \emph{validates} its
conclusion by verifying with the user.

In this ongoing work, if the explanation synthesizer identifies that the feedback is related 
to the user preferences, the Preference Reasoner will construct a new
case that is used by AToM to update the model of the user.  Even a single
piece of feedback from the user can be sufficient for AToM to learn the 
user's preferences because analogical learning, which is used by AToM, is
data efficient and capable of learning from only a few examples 
\cite{chen2019human,wilson2019analogical}.


\section{Related Work}
Consideration of \textit{user preferences} is an important aspect of human-robot interaction, as it allows the robot to modify its behaviors according to its understanding of the user. Hiatt, Harrison, and Trafton (\citeyear{hiatt2011accommodating}) found that people prefer collaborating with robots that adapt their behaviors in this way. However, most approaches to recognizing users' preferences use statistical techniques like reinforcement learning \cite{woodworth2018preference} and Markov Decision Processes \cite{munzer2017preference} to predict preferences. This means that they require large amounts of training data, are not responsive to user feedback, and are not explainable. That is, once trained, such systems predict preferences based on their built-up statistical models; a user cannot state a preference directly or inspect why the robot predicts a particular preference. By incorporating stated user preferences, and moving toward learning preferences by analogy, we attempt to avoid these pitfalls.

There are many forms of \textit{adaptive assistance} in robotics.  One approach is 
shared autonomy, in which the system infers human intentions and adapts how
much assistance in provided in controlling a robot \cite{Nikolaidis2017,Jain2019}.
This work is focused on assisting people in physically controlling robots, 
whereas we are working towards an autonomous robot that provides social assistance.

Other work has looked at adapting a robot's behavior based on user preferences.
For example, a recursive neural network was used
to learn weights pertaining to user preferences, which influence the plans for
a robot \cite{Bacciu2014}.  While the preferences did affect the plans used by a robot, 
the plans are used to improve the 
robot's navigation.  Thus, the user preferences do not relate to assistance 
provided to the user.

A model of Theory of Mind (ToM) has been proposed to adapt the assistance provided
by a social robot \cite{Gorur2017}.  A stochastic model is used to infer what action a person could 
be executing.  Based on this estimate, they generate a plan to determine with which 
action the robot should help. 
While they use ToM to estimate a user's intent (via a set of possible actions), 
they do not represent a user's preference for how the task should be completed.

  
One way to understand complex decision making systems is with
\emph{interpretable} or \emph{explainable} parts.  Explanations can describe
proxy methods \cite{why-trust,grad-cam,visualizing}, representations
\cite{netdissect2017,cavs}, or be inherently explanation-producing
\cite{multimodal}.  
In the context
of human-robot interaction, explanations can help to communicate and
build trust \cite{wang2016trust}, justify the robot's actions
\cite{stange2020effects} or motions \cite{dragan2013legibility}, or
describe \emph{unreasonable} perceptions \cite{gilpin-hri}.  But most of
these explanations are generated \emph{after-the-fact} and cannot be used
to improve the completion of tasks moving forward.  

We propose to use explanations as \emph{feedback} to augment assistive
robots.  This has been explored for agents playing games, especially
\emph{when} to provide explanations \cite{li2020reasoning}.  This approach
builds on Rainbow, a self-adaptive system that can correct itself and
reuse the same baseline framework \cite{rainbow}.  To our knowledge,
this is the first work to propose a knowledge-driven architecture that could use explanations to \emph{improve} robotic reasoning and inference. 
\section{Contributions}

In this paper, we motivate a knowledge-driven architecture for adaptive assistance. We demonstrate the functionality of the components of this architecture in a task for socially assistive robots (SARs). In future work, we will expand the architecture to incorporate and process feedback and learn user preferences. This paper opens a new area of research in adaptable and interpretable SARs.

\bibliographystyle{aaai}
\bibliography{pgaef}

\begin{thebibliography}{}

\bibitem[\protect\citeauthoryear{Bacciu \bgroup et al\mbox.\egroup
  }{2014}]{Bacciu2014}
Bacciu, D.; Gallicchio, C.; Micheli, A.; {Di Rocco}, M.; and Saffiotti, A.
\newblock 2014.
\newblock {Learning context-aware mobile robot navigation in home
  environments}.
\newblock In {\em IISA 2014 - 5th International Conference on Information,
  Intelligence, Systems and Applications},  57--62.

\bibitem[\protect\citeauthoryear{Bau \bgroup et al\mbox.\egroup
  }{2017}]{netdissect2017}
Bau, D.; Zhou, B.; Khosla, A.; Oliva, A.; and Torralba, A.
\newblock 2017.
\newblock Network dissection: Quantifying interpretability of deep visual
  representations.
\newblock In {\em Computer Vision and Pattern Recognition}.

\bibitem[\protect\citeauthoryear{Chen \bgroup et al\mbox.\egroup
  }{2019}]{chen2019human}
Chen, K.; Rabkina, I.; McLure, M.~D.; and Forbus, K.~D.
\newblock 2019.
\newblock Human-like sketch object recognition via analogical learning.
\newblock In {\em Proceedings of the AAAI Conference on Artificial
  Intelligence}, volume~33,  1336--1343.

\bibitem[\protect\citeauthoryear{Dragan, Lee, and
  Srinivasa}{2013}]{dragan2013legibility}
Dragan, A.~D.; Lee, K.~C.; and Srinivasa, S.~S.
\newblock 2013.
\newblock Legibility and predictability of robot motion.
\newblock In {\em 2013 8th ACM/IEEE International Conference on Human-Robot
  Interaction (HRI)},  301--308.
\newblock IEEE.

\bibitem[\protect\citeauthoryear{{Garlan} \bgroup et al\mbox.\egroup
  }{2004}]{rainbow}
{Garlan}, D.; {Cheng}, S.~.; {Huang}, A.~.; {Schmerl}, B.; and {Steenkiste}, P.
\newblock 2004.
\newblock Rainbow: architecture-based self-adaptation with reusable
  infrastructure.
\newblock {\em Computer} 37(10):46--54.

\bibitem[\protect\citeauthoryear{Gilpin \bgroup et al\mbox.\egroup
  }{2018}]{gilpin-hri}
Gilpin, L.~H.; Zaman, C.; Olson, D.; and Yuan, B.~Z.
\newblock 2018.
\newblock Reasonable perception: Connecting vision and language systems for
  validating scene descriptions.
\newblock In {\em Companion of the 2018 ACM/IEEE International Conference on
  Human-Robot Interaction},  115--116.

\bibitem[\protect\citeauthoryear{Gilpin, Macbeth, and
  Florentine}{2018}]{gilpin2018monitoring}
Gilpin, L.~H.; Macbeth, J.~C.; and Florentine, E.
\newblock 2018.
\newblock Monitoring scene understanders with conceptual primitive
  decomposition and commonsense knowledge.
\newblock {\em Advances in Cognitive Systems} 6:45--63.

\bibitem[\protect\citeauthoryear{Gilpin}{2020}]{leilanithesis}
Gilpin, L.~H.
\newblock 2020.
\newblock {\em Anomaly Detection through Explanations}.
\newblock Ph.D. Dissertation, Massachusetts Institute of Technology.

\bibitem[\protect\citeauthoryear{G{\"{o}}r{\"{u}}r, Rosman, and
  Hoffman}{2017}]{Gorur2017}
G{\"{o}}r{\"{u}}r, O.~C.; Rosman, B.; and Hoffman, G.
\newblock 2017.
\newblock {Toward Integrating Theory of Mind into Adaptive Decision- Making of
  Social Robots to Understand Human Intention}.
\newblock In {\em Workshop on the Role of Intentions in Human-Robot Interaction
  at the International Conference on Human-Robot Interaction}.

\bibitem[\protect\citeauthoryear{Hiatt, Harrison, and
  Trafton}{2011}]{hiatt2011accommodating}
Hiatt, L.~M.; Harrison, A.~M.; and Trafton, J.~G.
\newblock 2011.
\newblock Accommodating human variability in human-robot teams through theory
  of mind.
\newblock In {\em Twenty-Second International Joint Conference on Artificial
  Intelligence}.

\bibitem[\protect\citeauthoryear{Jain and Argall}{2019}]{Jain2019}
Jain, S., and Argall, B.
\newblock 2019.
\newblock {Probabilistic Human Intent Recognition for Shared Autonomy in
  Assistive Robotics}.
\newblock {\em ACM Transactions on Human-Robot Interaction (THRI)} 9(1):1--23.

\bibitem[\protect\citeauthoryear{Kim \bgroup et al\mbox.\egroup }{2017}]{cavs}
Kim, B.; Gilmer, J.; Viegas, F.; Erlingsson, U.; and Wattenberg, M.
\newblock 2017.
\newblock Tcav: Relative concept importance testing with linear concept
  activation vectors.
\newblock {\em arXiv preprint arXiv:1711.11279}.

\bibitem[\protect\citeauthoryear{Li \bgroup et al\mbox.\egroup
  }{2020}]{li2020reasoning}
Li, N.; C{\'a}mara, J.; Garlan, D.; and Schmerl, B.
\newblock 2020.
\newblock Reasoning about when to provide explanation for human-in-the-loop
  self-adaptive systems.

\bibitem[\protect\citeauthoryear{Munzer, Toussaint, and
  Lopes}{2017}]{munzer2017preference}
Munzer, T.; Toussaint, M.; and Lopes, M.
\newblock 2017.
\newblock Preference learning on the execution of collaborative human-robot
  tasks.
\newblock In {\em 2017 IEEE International Conference on Robotics and Automation
  (ICRA)},  879--885.
\newblock IEEE.

\bibitem[\protect\citeauthoryear{Nau \bgroup et al\mbox.\egroup
  }{1999}]{nau1999shop}
Nau, D.; Cao, Y.; Lotem, A.; and Munoz-Avila, H.
\newblock 1999.
\newblock Shop: Simple hierarchical ordered planner.
\newblock In {\em Proceedings of the 16th international joint conference on
  Artificial intelligence-Volume 2},  968--973.

\bibitem[\protect\citeauthoryear{Nikolaidis \bgroup et al\mbox.\egroup
  }{2017}]{Nikolaidis2017}
Nikolaidis, S.; Hsu, D.; Zhu, Y.~X.; and Srinivasa, S.
\newblock 2017.
\newblock {Human-Robot Mutual Adaptation in Shared Autonomy}.
\newblock In {\em 12th ACM/IEEE International Conference on Human-Robot
  Interaction (HRI)},  294--302.

\bibitem[\protect\citeauthoryear{Park \bgroup et al\mbox.\egroup
  }{2018}]{multimodal}
Park, D.~H.; Hendricks, L.~A.; Akata, Z.; Rohrbach, A.; Schiele, B.; Darrell,
  T.; and Rohrbach, M.
\newblock 2018.
\newblock Multimodal explanations: Justifying decisions and pointing to the
  evidence.
\newblock {\em CoRR} abs/1802.08129.

\bibitem[\protect\citeauthoryear{Rabkina and
  Forbus}{2019}]{rabkina2019analogical}
Rabkina, I., and Forbus, K.~D.
\newblock 2019.
\newblock Analogical reasoning for intent recognition and action prediction in
  multi-agent systems.
\newblock In {\em Proceedings of the Seventh Annual Conference on Advances in
  Cognitive Systems}.

\bibitem[\protect\citeauthoryear{Rabkina \bgroup et al\mbox.\egroup
  }{2017}]{rabkina2017towards}
Rabkina, I.; McFate, C.; Forbus, K.~D.; and Hoyos, C.
\newblock 2017.
\newblock Towards a computational analogical theory of mind.
\newblock In {\em Proceedings of the 39th Annual Meeting of the Cognitive
  Science Society}.

\bibitem[\protect\citeauthoryear{Rabkina \bgroup et al\mbox.\egroup
  }{2020}]{rabkina2020acs}
Rabkina, I.; Kantharaju, P.; Wilson, J.; Roberts, M.; Forbus, K.; and Hiatt,
  L.~M.
\newblock 2020.
\newblock Recognizing the goals of uninspecable agents.
\newblock {\em Proceedings of the Eighth Annual Conference on Advances in
  Cognitive Systems}.

\bibitem[\protect\citeauthoryear{Rabkina, McFate, and
  Forbus}{2018}]{rabkina2018bootstrapping}
Rabkina, I.; McFate, C.; and Forbus, K.~D.
\newblock 2018.
\newblock Bootstrapping from language in the analogical theory of mind model.
\newblock In {\em Proceedings of the 40th Annual Meeting of the Cognitive
  Science Society}.

\bibitem[\protect\citeauthoryear{Ribeiro, Singh, and
  Guestrin}{2016}]{why-trust}
Ribeiro, M.~T.; Singh, S.; and Guestrin, C.
\newblock 2016.
\newblock Why should i trust you?: Explaining the predictions of any
  classifier.
\newblock In {\em Proceedings of the 22nd ACM SIGKDD International Conference
  on Knowledge Discovery and Data Mining},  1135--1144.
\newblock ACM.

\bibitem[\protect\citeauthoryear{Selvaraju \bgroup et al\mbox.\egroup
  }{2016}]{grad-cam}
Selvaraju, R.~R.; Cogswell, M.; Das, A.; Vedantam, R.; Parikh, D.; and Batra,
  D.
\newblock 2016.
\newblock Grad-cam: Visual explanations from deep networks via gradient-based
  localization.
\newblock {\em See https://arxiv. org/abs/1610.02391 v3} 7(8).

\bibitem[\protect\citeauthoryear{Stange and Kopp}{2020}]{stange2020effects}
Stange, S., and Kopp, S.
\newblock 2020.
\newblock Effects of a social robot's self-explanations on how humans
  understand and evaluate its behavior.
\newblock In {\em Proceedings of the 2020 ACM/IEEE International Conference on
  Human-Robot Interaction},  619--627.

\bibitem[\protect\citeauthoryear{Wang, Pynadath, and
  Hill}{2016}]{wang2016trust}
Wang, N.; Pynadath, D.~V.; and Hill, S.~G.
\newblock 2016.
\newblock Trust calibration within a human-robot team: Comparing automatically
  generated explanations.
\newblock In {\em 2016 11th ACM/IEEE International Conference on Human-Robot
  Interaction (HRI)},  109--116.
\newblock IEEE.

\bibitem[\protect\citeauthoryear{Wilson \bgroup et al\mbox.\egroup
  }{2019a}]{wilson2019analogical}
Wilson, J.~R.; Chen, K.; Crouse, M.; Nakos, C.; Ribeiro, D.~N.; Rabkina, I.;
  and Forbus, K.~D.
\newblock 2019a.
\newblock Analogical question answering in a multimodal information kiosk.
\newblock In {\em Proceedings of the Seventh Annual Conference on Advances in
  Cognitive Systems}.

\bibitem[\protect\citeauthoryear{Wilson \bgroup et al\mbox.\egroup
  }{2019b}]{wilson2019developing}
Wilson, J.~R.; Kim, S.; Kurylo, U.; Cummings, J.; and Tarneja, E.
\newblock 2019b.
\newblock Developing computational models of social assistance to guide
  socially assistive robots.
\newblock In {\em Proceedings of the AI-HRI Symposium at AAAI-FSS 2019}.

\bibitem[\protect\citeauthoryear{Wilson, Tickle-Degnen, and
  Scheutz}{2020}]{wilson2020challenges}
Wilson, J.~R.; Tickle-Degnen, L.; and Scheutz, M.
\newblock 2020.
\newblock Challenges in designing a fully autonomous socially assistive robot
  for people with parkinson’s disease.
\newblock {\em ACM Transactions on Human-Robot Interaction (THRI)} 9(3):1--31.

\bibitem[\protect\citeauthoryear{Wilson, Wransky, and
  Tierno}{2018}]{wilson2018general}
Wilson, J.~R.; Wransky, M.; and Tierno, J.
\newblock 2018.
\newblock General approach to automatically generating need-based assistance.
\newblock In {\em Proceedings of the Sixth Annual Conference on Advances in
  Cognitive Systems}.

\bibitem[\protect\citeauthoryear{Woodworth \bgroup et al\mbox.\egroup
  }{2018}]{woodworth2018preference}
Woodworth, B.; Ferrari, F.; Zosa, T.~E.; and Riek, L.~D.
\newblock 2018.
\newblock Preference learning in assistive robotics: Observational repeated
  inverse reinforcement learning.
\newblock In {\em Machine Learning for Healthcare Conference},  420--439.

\bibitem[\protect\citeauthoryear{Zeiler and Fergus}{2014}]{visualizing}
Zeiler, M.~D., and Fergus, R.
\newblock 2014.
\newblock Visualizing and understanding convolutional networks.
\newblock In {\em European conference on computer vision},  818--833.
\newblock Springer.

\end{thebibliography}
\end{document}